\documentclass[journal]{IEEEtran}

\usepackage{graphicx}
\usepackage{comment}
\usepackage{amsmath,amssymb} 
\usepackage{color}

\usepackage{multirow}
\usepackage[ruled]{algorithm2e}
\usepackage{booktabs}
\usepackage{subfigure}

\ifCLASSINFOpdf
\else
\fi
\begin{document}
%
\title{Dual ResGCN for Balanced Scene Graph Generation}
%
%
%

\author{Jingyi Zhang,
        Yong Zhang,
        Baoyuan Wu, ~\IEEEmembership{Member,~IEEE,}
        Yanbo Fan, 
        Fumin Shen, ~\IEEEmembership{Member,~IEEE,} \\
        and~Heng Tao Shen, ~\IEEEmembership{Senior Member,~IEEE,} 
\thanks{Yong Zhang and Baoyuan Wu are corresponding authors.}
\thanks{J. Zhang, F. Shen, and H. T. Shen are with the Center for
Future Media, University of Electronic Science and Technology of China,
Chengdu 610054, China, and also with the School of Computer Science
and Engineering, University of Electronic Science and Technology of China,
Chengdu 610054, China (e-mail: jingyi.zhang1995@gmail.com; fumin.shen@gmail.com;  shenhengtao@hotmail.com).).}
\thanks{B. Wu is with the School of Data Science, the Chinese University of Hong Kong, Shenzhen, Shenzhen, 518172 China, and also with the Secure Computing Lab of Big Data, Shenzhen Research Institute of Big Data,   Shenzhen, 518172 China (e-mail: wubaoyuan1987@gmail.com).}
\thanks{Y. Zhang and Y. Fan are with the Tencent AI Lab, Shenzhen 519000,
China (e-mail: zhangyong201303@gmail.com; fanyanbo0124@gmail.com).}
\thanks{This work was done when Jingyi Zhang was an intern at Tencent AI Lab.}
}

%
%

\markboth{IEEE TRANSACTIONS ON IMAGE PROCESSING, VOL. XX, NO. XX, SEPTEMBER, 2020}%
{Shell \MakeLowercase{\textit{et al.}}: Bare Demo of IEEEtran.cls for IEEE Journals}
%



\maketitle

\begin{abstract}
  Visual scene graph generation is a challenging task. Previous works have achieved great progress, but most of them do not explicitly consider the class imbalance issue in scene graph generation. 
  Models learned without considering the class imbalance tend to predict the majority classes, which leads to a good performance on trivial frequent predicates, but poor performance on informative infrequent predicates.
  However, predicates of minority classes often carry more semantic and precise information~(\textit{e.g.}, \emph{`on'} v.s \emph{`parked on'}).
  To alleviate the influence of the class imbalance, we propose a novel model, dubbed \textit{dual ResGCN}, which consists of an object residual graph convolutional network and a relation residual graph convolutional network. 
  The two networks are complementary to each other. 
  The former captures object-level context information, \textit{i.e.,} the connections among objects. We propose a novel ResGCN that enhances object features in a cross attention manner. Besides, we stack multiple contextual coefficients to alleviate the imbalance issue and enrich the prediction diversity.
  The latter is carefully designed to explicitly capture relation-level context information \textit{i.e.,} the connections among relations. We propose to incorporate the prior about the co-occurrence of relation pairs into the graph to further help alleviate the class imbalance issue. 
  Extensive evaluations of three tasks are performed on the large-scale database VG to demonstrate the superiority of the proposed method.  
\end{abstract}

\begin{IEEEkeywords}
Scene Graph Generation, Class Imbalance, Dual Residual Graph Convolutional Network
\end{IEEEkeywords}

\IEEEpeerreviewmaketitle

\section{Introduction}


\IEEEPARstart{V}{isual} scene graph generation (SGG) aims to identify the classes of objects and their relations in a given image, which is a challenging vision task. 
Different from other tasks such as object detection~\cite{ssd,yolo9000}, object classification~\cite{resnet,DBLP:conf/eccv/DengDJFMBLNA14} and semantic segmentation~\cite{DBLP:conf/cvpr/LongSD15,DBLP:journals/pami/ChenPKMY18}, scene understanding goes beyond the whereabouts of objects, but more importantly establishes a scene graph among objects through the transfer and mutual understanding of relations. 
Such a scene graph provides a structured representation of an image that can support a wide range of high-level visual tasks, including image captioning~\cite{DBLP:conf/eccv/GuJCW18}, visual question answering~\cite{DBLP:journals/pami/WuSWDH18,DBLP:conf/icml/XiongMS16}, image retrieval \cite{wu2018multi,DBLP:conf/cvpr/JohnsonKSLSBL15}, and image generation~\cite{DBLP:conf/cvpr/JohnsonGF18}.

\begin{figure*}[t]
  \centering
  \subfigure{
  \begin{minipage}[t]{0.5\linewidth}
  \centering
  \includegraphics[width = 1.0\columnwidth]{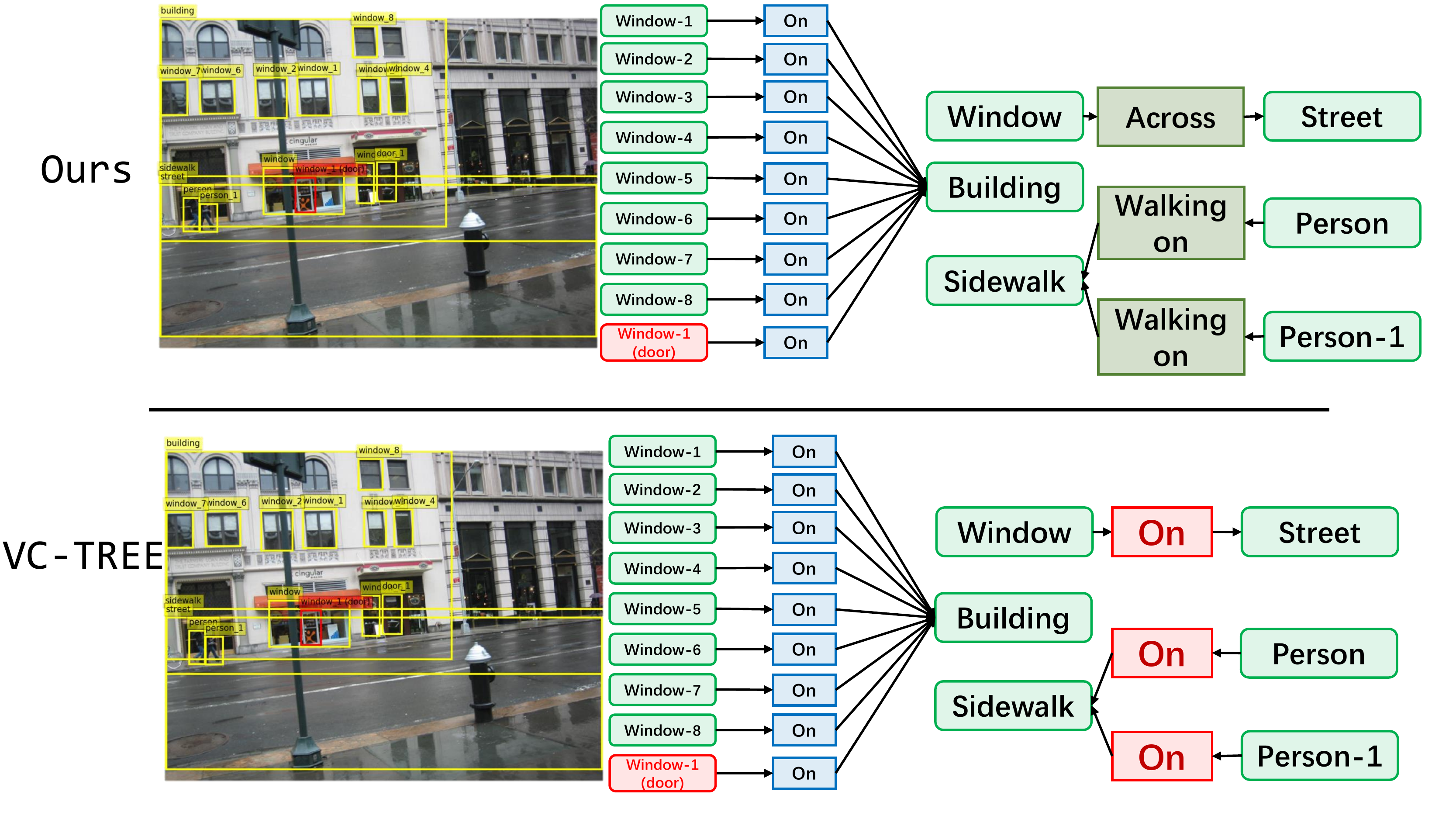}
  \end{minipage}%
  }%
  \subfigure{
  \begin{minipage}[t]{0.5\linewidth}
  \centering
  \includegraphics[width = 1.0\columnwidth]{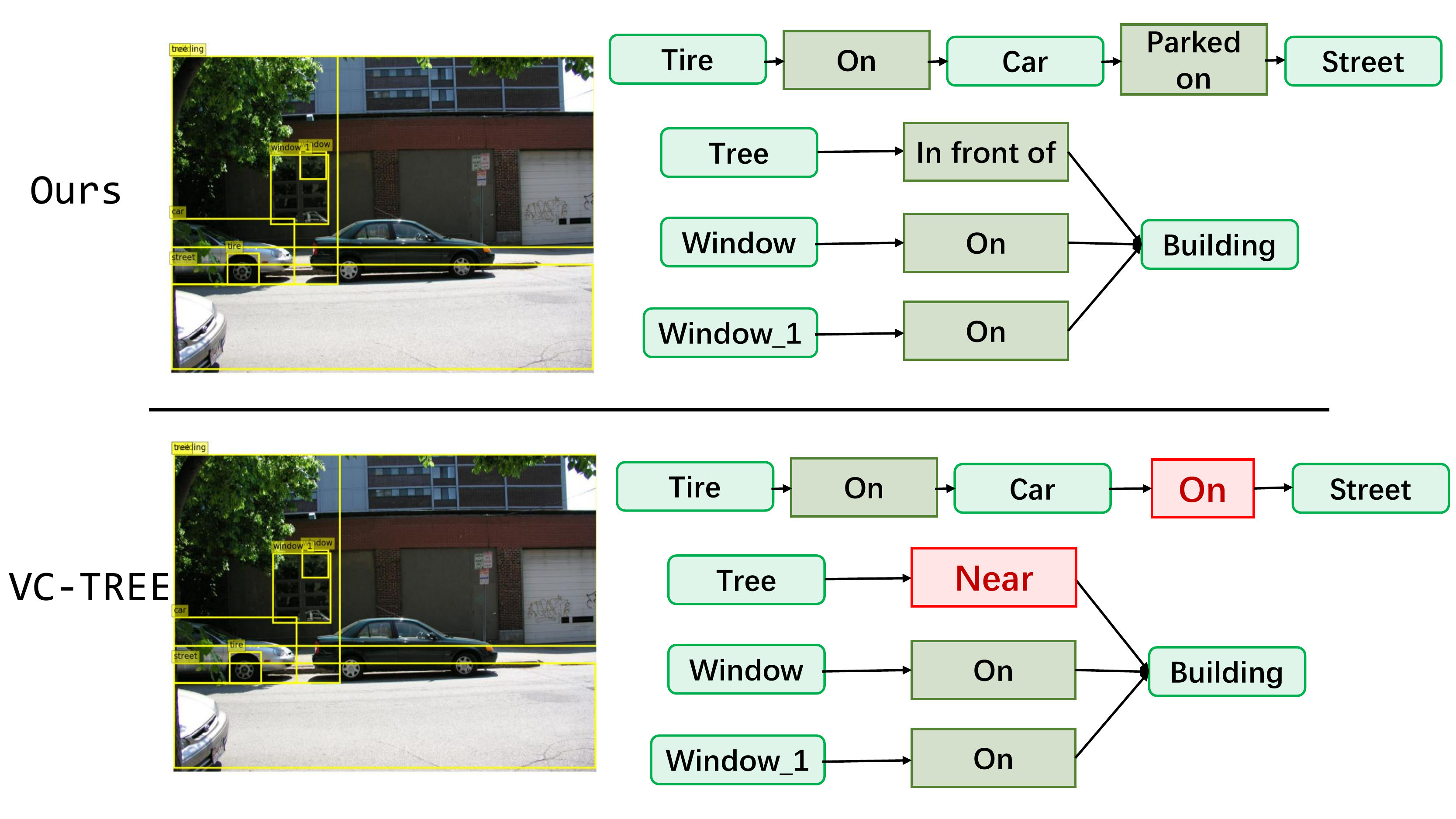}
  \end{minipage}%
  }%
  \centering
  
  \caption{Illustration of generated scene graphs. The upper part illustrates the results of our method. The lower part illustrates the results of SOTA method VC-TREE \cite{vctree}. Red boxes indicate false predictions. Compared to VC-TREE, our method that successfully predicts the infrequent relations gives a more in-depth and more fine-grained understanding of the complicated scene.}
  \label{fig:showresults}
  
  \end{figure*}

In existing works, there are two categories of methods
for scene graph generation. Methods in the first category jointly predict the classes of objects and their relations based on the object and relation proposals~\cite{LiOWT17,factornet,DBLP:journals/corr/abs-1904-02104,GuZL0CL19}. Methods in the second category detect the objects first and then predict the relations of object pairs~\cite{kern,DBLP:journals/corr/abs-1812-02347,vctree,linknet,motif}. 
Among these methods, the chain graph~\cite{motif} and the fully connected graphs~\cite{XuZCF17,DBLP:journals/corr/abs-1904-02104,graphrcnn,factornet,DBLP:journals/corr/abs-1812-02347} are two prevalent types of structures to model the relations among objects, which have made great progress for scene graph generation.

However, these methods are developed without the explicit consideration of the class imbalance problem in the existing databases of scene graph generation.
As revealed in~\cite{vctree}, some relations such as `on' occupy a large portion in the databases, referred to as \textit{frequent relations}, while some other relations such as `parked on' and `walking in' occupy a small portion, referred to as \textit{infrequent relations}. If a model is directly trained on a database whose class distribution is seriously skewed, it tends to predict a relation to be a frequent relation rather than an infrequent relation. 
For example, the bottom row of Fig.~\ref{fig:showresults} shows the scene graph generated by a state-of-the-art model~\cite{vctree}. 
All the predicted relations are frequent relations, \textit{i.e.}, `on' or `near', but the infrequent relations such as `parked on' and `in front of' are more accurate and provide more fine-grained understanding for complicate interactions among objects. 

In terms of quantitative performance, learning a model without explicitly considering the class imbalance will lead to a good score in Recall, but a poor score in mean Recall. 
Because when a model tends to predict the frequent relations, the recall would be high for these frequent relations, but the recall could be low for infrequent ones. As frequent relations dominate the portion of the whole database, the total Recall is high, but the mean Recall is poor because of the poor performance of infrequent relations. 
Hence, both Recall and mean Recall are important and necessary evaluation metrics when the database is seriously skewed, especially the mean recall.

To alleviate the influence of the class imbalance on the prediction of infrequent relations, we propose a novel network, \textit{i.e., dual ResGCN}, which consists of an object residual graph convolutional network (\textit{Object ResGCN}) and a relation residual graph convolutional network (\textit{Relation ResGCN}). They are complementary to each other. 
The former takes object features as nodes and aggregates object-level context information. Different previous methods which utilize the popular first-order linear model, 
we adopt a novel cross-attention model to produce attention
maps that guide feature aggregation.
Besides, we stack multiple contextual coefficients to help alleviate the imbalance issue and enrich the prediction diversity.
The latter treats relation features as nodes and aggregates relation-level context information among relations. It explicitly models the connections among relations to help alleviate the class imbalance issue.  
As revealed in~\cite{DBLP:conf/cvpr/ShiZCL19a}, the second-order features in graph (\textit{i.e.}, the edge features) are more informative and discriminative. 
It captures higher-level information and outputs more discriminative features.
If using only the \textit{Object ResGCN}, the prediction of the relation depends on the information from the mounted objects. 
It tends to predict the frequent relation according to the co-occurrence of two mounted objects. 
On the contrary, when cooperating with the complementary \textit{Relation ResGCN}, the prediction of the relation depends on not only the information from the associated objects but also the information from other relation nodes.  
Moreover, the inputs of the \textit{Relation ResGCN} consist of the features of the objects as well as the visual features of their union region. 
The union region contains the contextual information of the interacted objects, which provides some cues to help alleviate the influence of the class imbalance.


Besides the novel network, we propose to incorporate prior knowledge about relation pairs into the graph to alleviate the class imbalance issue. 
They are exploited in the context aggregation process for the \textit{Relation ResGCN}. 
Good context prior ( \textit{e.g.,} \emph{man wearing shirt} rather than \emph{riding}) can effectively regularize the distribution of possible relations of object pairs and thus makes prediction less ambiguous~\cite{tang2020unbiased}, while the undesirable long-tailed prior makes the relation prediction collapse (\textit{e.g.}, \emph{parked on, sitting on} collapsed to \emph{on}).
To take advantage of the good prior and avoid the long-tailed prior, we count the co-occurrence probability of all relations. 
To alleviate the side effects of the long-tailed bias, we count a type of relation only once in an image, \textit{i.e.}, if a relation occurs multiple times in an image, we count it only once.
Combining the learnable graph with the prior graph can effectively regularize the graph structure, leading to more balanced predictions.
Please note that using the class distribution to weigh the losses in the objective is a common strategy to tackle the class imbalance. 
However, this strategy introduces a set of hyperparameters to tune. 
Differently, we incorporate the prior into the structure without additional hyperparameters.

The main contributions of this work are two-fold.
\begin{itemize} 
\item We propose a novel and effective complementary network, dubbed \textit{dual ResGCN}, which alleviates the class imbalance issue by capturing the relation-level and object-level contexts, as well as the co-occurrence prior of relation. 
\item  We perform extensive experiments and ablation studies of three tasks (SGDet, SGCls, and PredCls) on the large-scale dataset VG to demonstrate the superiority of our method. 
 Our absolute improvement of infrequent relations over state-of-the-art methods is up to 17.74\%.

\end{itemize}

\section{Related Work}
\subsection{Scene Graph Generation} The notion of the scene graph is developed from visual relation detection. As a pioneer work, Lu \emph{et al.}~\cite{LuKBL16} introduced generic visual relation detection as a visual task.
In the early stage, many works~\cite{LuKBL16,ZhangKCC17,ZhangECCE17,ZhuJ18,ZhuangLS017} regard objects as isolated individuals and recognize relations between those independent object pairs. Nevertheless, these models overlook the importance of visual context. To benefit from rich global context, recent works consider each image as a whole and utilize the message passing 
mechanism~\cite{motif,kern,vctree,DBLP:journals/corr/abs-1904-02104,XuZCF17}. 
\cite{XuZCF17} introduced an end-to-end model
that learns to iteratively refine relationships and object predictions via message passing based on the RNNs. \cite{DBLP:conf/iccv/LiOZWW17} proposed a multi-task framework to utilize 
semantic associations over three tasks of object detection,
scene graph generation, and image captioning. They found that jointly learning the three tasks achieves better performance.
Dai \textit{et al.}  
 \cite{motif}  firstly
brought the bias problem of SGG into attention. It presents an analysis of statistical co-occurrences between relationships and object pairs on the Visual Genome
dataset~\cite{DBLP:journals/ijcv/KrishnaZGJHKCKL17} and draws a conclusion that these statistical co-occurrence priors provids strong regularization for relationship prediction. Besides, it encodes the global context of objects and relationships by LSTMs to
facilitate scene graph parsing. \cite{kern} incorporated statistical correlations into deep neural networks.
 \cite{vctree} utilized a dynamic tree structure to further capture the intrinsic property of scene graph and refine the results.  Although~\cite{kern} and~\cite{motif} proposed an unbiased metric (\textit{i.e.}, mean Recall), yet their performance on infrequent relationships is still far from satisfactory.

The most related work to ours is \cite{kern}. They built a graph representation of statistical correlations between object pairs and their relationships and used a graph neural network to learn the interaction between relationships and objects to generate a scene graph.  Our work differs from~\cite{kern} in two aspects. (1) Different priors. We propose a more balanced relation prior for the first time. (2) Graph of~\cite{kern} with fixed edge weights may ignore coherent relations among visual features and lacks flexibility. In contrast, our graph structure is a combination of a learnable graph with a prior graph, which can effectively regularize the graph structure, leading to more balanced predictions.

\subsection{Class Imbalance in Scene Graph Generation}
Alleviating the class imbalance problem in the training process is crucial to achieve an optimal training and fully exploit the potential of model architecture. It has been well-studied in many visual tasks like image recognition~\cite{DBLP:conf/cvpr/HuangLLT16}, segmentation~\cite{DBLP:journals/corr/WuSH16a}, and object detection~\cite{DBLP:conf/cvpr/PangCSFOL19}. 
However, to the best of our knowledge, only a few recent works have explicitly tackled the class imbalance problem in SGG. 
\cite{motif} firstly pointed out the imbalance problem.
\cite{vctree} and \cite{kern} proposed a new balanced evaluation metric, \textit{i.e.}, mean Recall.
Besides, \cite{kern} unified the statistical knowledge with a deep architecture to facilitate scene graph generation. \cite{GuZL0CL19} exploited external knowledge and image reconstruction to refine image features. \cite{vrvg} pruned those easy frequent relations and introduced a new dataset. 
However, the performance of those methods is far from satisfactory.

\subsection{Residual Graph Convolutional Network} Recent works~\cite{KipfW17,LiHW18,XuZCF17} showed that stacking multiple layers of graph convolutions leads to high complexity in back-propagation and their performance degrades when the number of layers is greater than 3. 
\cite{BaldwinCR18} proposed a Highway GCN which adds a ``highway" gate between layers to avoid vanishing gradients. \cite{DBLP:journals/corr/abs-1904-03751} borrowed the ideas from ResNet and introduced ResGCN which adds skip connections between layers. Combined with Dynamic Edges with dilation, ResGCN can be very deep and achieves promising performance.

\section{The Proposed Approach}

\begin{figure}[t]
    \centering
    \includegraphics[width = 0.5\textwidth]{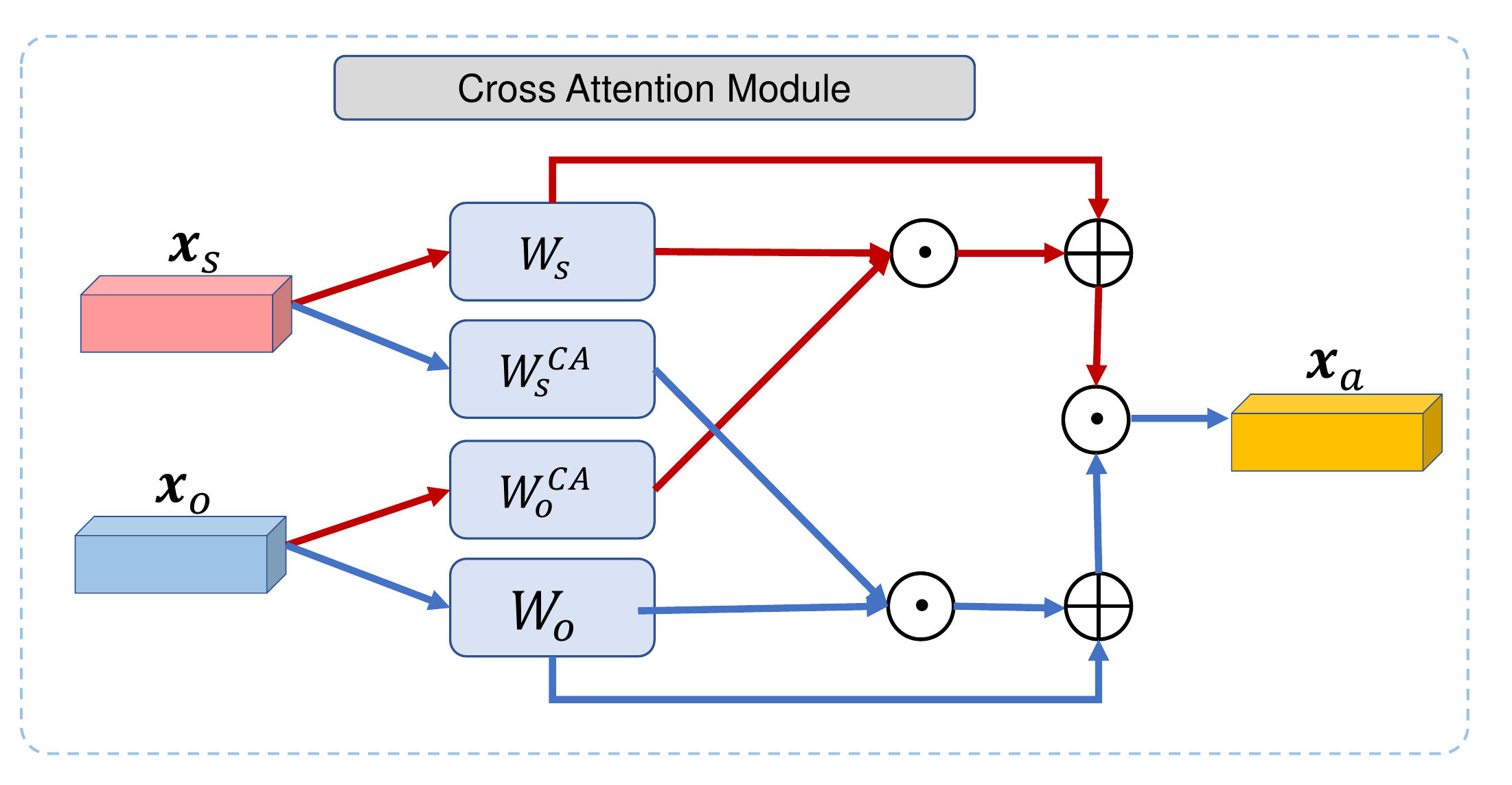}
    \caption{Illustration of our Cross-attention ($\mathcal{CA}$) module. $\otimes$ and $\oplus$ denote element-wise product and sum, respectively.}
    \label{fig:crossAtten}
\end{figure}

\noindent \textbf{{Formal Definition.}} Given a set of predefined object classes $\mathcal{C}$ (including the
background) and a set of visual relationship classes $\mathcal{R}$ (including
non-relationship), a scene graph is formally represented by $\mathcal{G} = \{\mathcal{V} = \{\boldsymbol{v_i}, \boldsymbol{b_i}\},\mathcal{E} = \{ \boldsymbol{r_{ij}} \} \}$, where the $\mathcal{G}$ and $\mathcal{E}$ denote the set of nodes and the set of edges, respectively. $\boldsymbol{v_i} \in \mathcal{C}$ denotes the object class of the $i$-th object node, while $\boldsymbol{b_i} \in \mathbb{R}^4$ is the bounding box of the $i$-th object. $\boldsymbol{r_{ij}} \in \mathcal{R} $ denotes the relationship between the $i$-th and  the $j$-th object node. Scene Graph Generation (SGG) is to formulate a structured representation of the semantic message in an image.

\begin{figure*}[t]
    \centering
    \includegraphics[width = 0.99\textwidth]{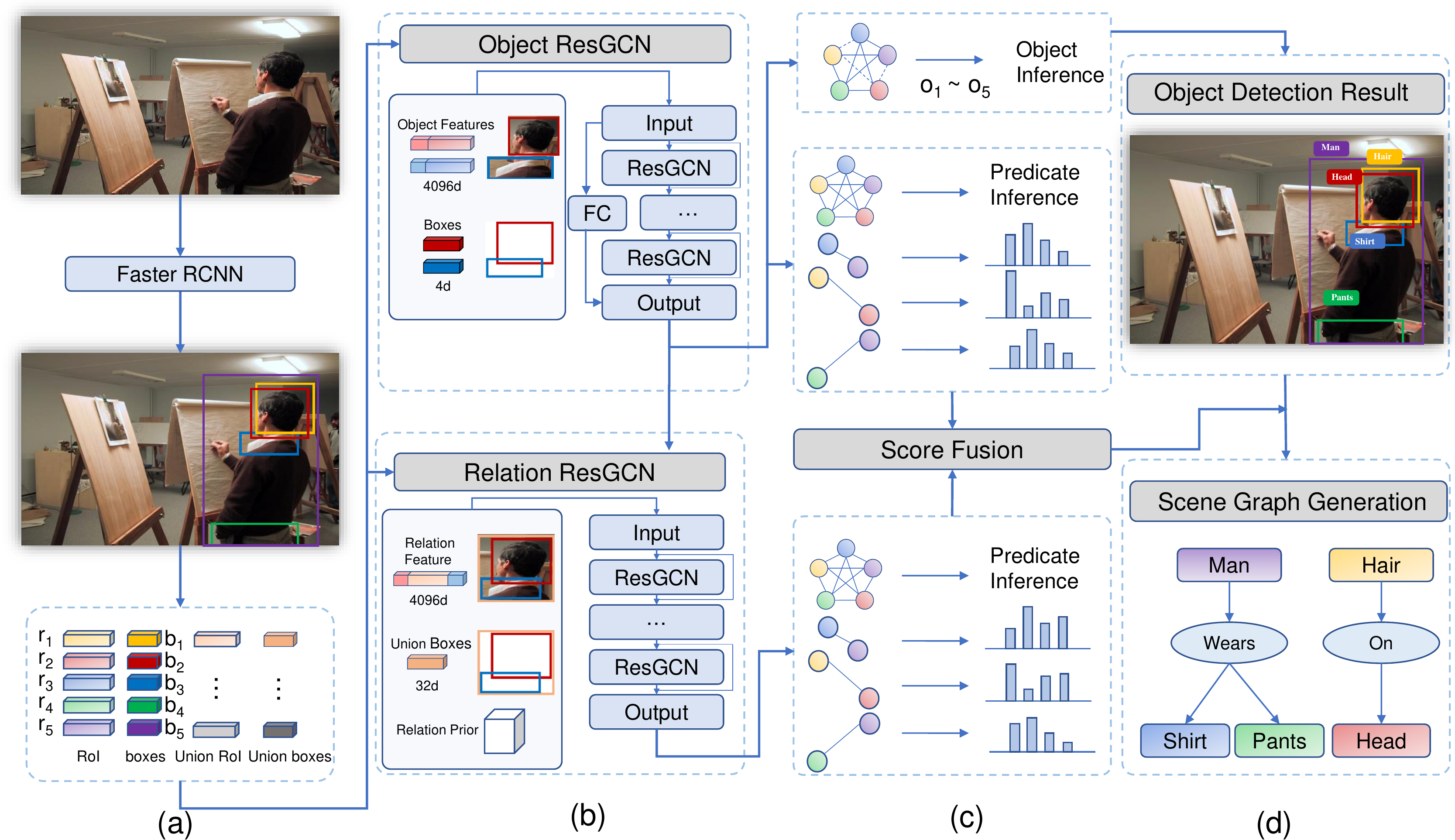}
    \caption{Overview of our framework. (a) Feature Extraction: we adopt Faster RCNN to extract ROI features.  (b) \textit{dual ResGCN}: we build two separate graphs for objects and relations, and employ an \textit{Object ResGCN} and a  \textit{Relation ResGCN} to obtain the object/relation class logits. (c) Score Fusion:  We fuse all the logits by element-wise summation and infer their relations. 
    (d) Scene Graph Generation: The process is repeated for all the object pairs, and the scene graph is generated.}
    \label{fig:Overview}
\end{figure*}

As illustrated in Fig. \ref{fig:Overview}, our framework for scene graph generation can be summarized as the following four steps. First, we adopt Faster R-CNN~\cite{DBLP:journals/pami/RenHG017} with the VGG backbone to detect object proposals. We adopt the same way as~\cite{motif} to obtain the features of each proposal. For each object region, the visual features are formulated by concatenating the appearance features $\boldsymbol{f}^o_i$, the bounding box $\boldsymbol{b}_i $, and the object classification confidence scores $\boldsymbol{s}_i$. Then, the concatenated features are projected into a lower-dimensional subspace and denoted as $\boldsymbol{x}_i$. Besides, we extract the visual features $\boldsymbol{u}_{ij}$ of the $i$-th and the $j$-th detected region.
These feature vectors are then fed into the propagation networks for the subsequent inference. 
Second, an object ResGCN with cross attention is exploited to aggregate object-level context and predict the object labels. 
Third, a relation ResGCN with relation prior is used to encode relation-level contextual information. 
Finally, based on the outputs of the object ResGCN and the relation ResGCN, the scene graph is generated.

We first introduce the proposed \textit{dual ResGCN} which consists of an \textit{Object ResGCN} and a \textit{Relation ResGCN}. 
As mentioned before, the class distributions of existing databases are seriously skewed. For instance, the most frequent relations (\emph{`on', `wearing', `has', `of', and `in'})  occupy 76\% portion in the databases. 
Besides, most methods~\cite{motif,factornet,GuZL0CL19,vctree} tend to achieve relation context features from object pairs. 
However, given the categories of a pair of objects, the probability distribution of their predicate is highly skewed for these typical frequent relations~\cite{motif,kern}. 
As a result, the predictions of the model are dominated by frequent relations with such large prior probabilities. To alleviate the class imbalance issue, we build two separate branches to aggregate context features from object embedding as well as relation embedding. Inspired by the skeleton-based action recognition method~\cite{skelton1},  we build a two-stream architecture to learn sufficient relational information from both objects and relations in an image. 
As edges (relation features) in a graph are naturally more informative and discriminative~\cite{DBLP:conf/cvpr/ShiZCL19a}, they emphasize the interactions between objects and relative positions, which provide higher-level information for relation prediction. While points (object features) emphasize the absolute positions and provide fundamental relation information by analyzing the distribution or local density of objects' visual features.

\subsection{Object ResGCN}
\label{section:3.1}
Our \textit{Object ResGCN} can be summarized as the following four steps.
(a) The Cross-attention module ($\mathcal{CA}$)  takes object features and union region features as inputs. Its output features denote the high-level interactions among inputs.
(b) We project the cross attention from (a) into contextual coefficients. Note that we compute multiple contextual coefficients for a given node to enrich the prediction diversity and alleviate the imbalance issue.
(c) Base on the contextual coefficients from (b), we refine the original node features through a residual connection. (d)  Finally, we use the refined features from (c) to achieve the object/relation predictions.

Existing methods of scene graph generation usually 
compute the  pairwise contextual coefficients between
subject and object by a dot product operation, but ignore the semantic relevance~\cite{GPS} and the rich contextual information of the union region~\cite{largekernel}.
For instance, if there are a man and multiple wine glasses in an image, previous methods are quite likely to make all biased predictions `\emph{holding}' according to the co-occurrence prior. However, they fail to determine exactly which wine glass is being held. We propose a Residual Cross-attention GCN (ResCAGCN) which can model the semantic relevance among the subject features, object features, and the union region features, thus draw attention to the target objects and benefit the subsequent feature aggregation and relation prediction.

\subsubsection{Cross-attention Feature Learning}
 Inspired by~\cite{crossAtten}, we propose our cross-attention module ($\mathcal{CA}$) to capture the semantic relevance.
The cross-attention module ($\mathcal{CA}$) is illustrated in Fig.~\ref{fig:crossAtten}. It takes node features $\boldsymbol{x}_i$ and $\boldsymbol{x}_j$ as inputs. We first design a correlation layer to calculate the cross-attention features of  $\boldsymbol{x}_i$ and $\boldsymbol{x}_j$, which are then used to calculate the contextual  coefficients and guide the aggregation of the node features. The cross-attention features are computed as:
\begin{equation}\label{eq:equiv1}
\begin{split}
\boldsymbol{x}^{a}_{ij} = \mathcal{CA}(\boldsymbol{x}_i, \boldsymbol{x}_j) =  (\boldsymbol{W}_s \boldsymbol{x}_i \odot \sigma( \boldsymbol{W}^{CA}_o \boldsymbol{x}_j) + \boldsymbol{W}_s \boldsymbol{x}_i) \\
\odot (\boldsymbol{W}_o \boldsymbol{x}_j \odot \sigma( \boldsymbol{W}^{CA}_s \boldsymbol{x}_i) + \boldsymbol{W}_o \boldsymbol{x}_j),
\end{split}
\end{equation}
where $\boldsymbol{\odot}$ represents Hadamard product. $\sigma$ is the Sigmod function to normalize the attention scores.
$\boldsymbol{W}_s$, $\boldsymbol{W}_o$, $\boldsymbol{W}^{CA}_s$, and $\boldsymbol{W}^{CA}_o$ are projection matrices for fusion. We use different transformation matrices to specify the positions of the subjects and the objects, which enables the function $\mathcal{CA}$ to be aware of the edge direction information. Besides, the cross attention is also a powerful manner to model the sophisticated interactions between input features.

\subsubsection{Contextual  Coefficient Learning} Given two node features $\boldsymbol{x}_i$ and $\boldsymbol{x}_j$, and union region features $\boldsymbol{u}_{ij}$, the contextual  coefficients $\boldsymbol{s}_{ij}$ are computed as follows:
\begin{equation}\label{eq:equiv2}
\begin{split}
\boldsymbol{s}_{ij} = \sigma( \boldsymbol{W}^{T}( \mathcal{CA}(\mathcal{CA}(\boldsymbol{x}_i, \boldsymbol{x}_i), \boldsymbol{u}_{ij}))),
\end{split}
\end{equation}
where $\mathcal{CA}$ denotes the cross-attention module. Note that: (1) Taking the union region features $\boldsymbol{u}_{ij}$ into consideration can enlarge the receptive field in context modeling.
(2) Most previous methods utilize the cosine similarity metric to measure the contextual coefficients, which may generate similar context information for all nodes and lead to structure  bias~\cite{GPS,tang2020unbiased}. On the contrary, 
we couple features together by Hadamard product and then project it to contextual coefficients, making the contextual information more correlated to specific regions.  
(3) To further alleviate the imbalance issue and enrich the prediction diversity, we borrow the idea from multi-head attention which computes multiple contextual coefficients for given nodes. Then, we stack them as vector $[\boldsymbol{s}^1_{ij}, \boldsymbol{s}^2_{ij}, \cdots , \boldsymbol{s}^N_{ij}]^T$.

\subsubsection{Feature Aggregation} We do not directly use the aggregated features as our output features. Instead, we use a residual connection to add them back to
the original features. The \textit{Object ResGCN} is formulated as:
\begin{equation}\label{eq:equiv3}
\begin{split}
\boldsymbol{\hat{x}}_{i} =  \boldsymbol{x}_{i} + ReLU( \boldsymbol{W}_3 LN(\boldsymbol{W}_2 \sum_{j \in \mathcal{N}_i} \boldsymbol{s}_{ij} \otimes \boldsymbol{W}_3 \boldsymbol{x}_j)),
\end{split}
\end{equation}
where $\otimes$ denotes Kronecker product. $\boldsymbol{W}_1, \boldsymbol{W}_2$, and $\boldsymbol{W}_3$ are linear transformations to embed features into same latent space. $\mathcal{N}_i$ denotes the $i$-th node's neighborhood. $LN$ denotes the layer normalization, which is utilized to filter the redundant information and refine the obtained contextual information~\cite{gcnet, GPS}. 

\subsubsection{Object Prediction} 
Finally, we feed the refined object context features $\boldsymbol{\hat{x}}_i$ into a classifier $\rho_o$ to predict the class label:
\begin{equation}
    P^{o}_i=\rho_o(\boldsymbol{\hat{x}}_i).\\
   \label{eq:object-cls}
\end{equation}
The predicted class label $C^p_{i}=\mathrm{argmax}(P^{o}_i)$ is then used for object inference.

\subsubsection{Relation Prediction} 
We collect pairwise features of each object pair: (1) $\boldsymbol{\hat{x}}_i$ and $\boldsymbol{\hat{x}}_j$ as contextual features and (2) $\boldsymbol{u}_{ij}$ as the visual features of the union region of the object pair. Then, we fuse them into the final features:
\begin{equation}
    \boldsymbol{r}_{ij}= \boldsymbol{\hat{x}}_i \oplus \boldsymbol{\hat{x}}_j \oplus \boldsymbol{u}_{ij}, \\
   \label{eq:relation-feat}
\end{equation}
where $\oplus$ denotes the fusion function defined in~\cite{vctree,nlpfusion}, \textit{i.e.,} $\boldsymbol{x} \oplus \boldsymbol{y} = ReLU(\boldsymbol{W}_x \boldsymbol{x} +\boldsymbol{W}_y \boldsymbol{y}) - (\boldsymbol{W}_x \boldsymbol{x} - \boldsymbol{W}_y \boldsymbol{y})\otimes (\boldsymbol{W}_x \boldsymbol{x} - \boldsymbol{W}_y \boldsymbol{y})$.
Then, we pass $\boldsymbol{r}_{ij}$ into a classifier to achieve distribution of the $k$-th relation label $P^{ro}_k$.
Additionally, based on the observation that many relations, such as human interactions, can be accurately inferred by the appearance of only the subjects or objects~\cite{loss}. 
We also introduce skip connections to attain the predicate class logits, which are only conditioned on the subject or object ROI features. Finally, we add these subject-only logits and object-only logits to $P^{ro}_k$ and get the final scores for the \textit{Object ResGCN} $P^{r\hat{o}}_k$.

\subsection{Relation ResGCN}
\label{section:3.2}

We define the relation region as the smallest rectangle which covers the union region of the $i$-th object and the $j$-th subject. And the $k$-th ($k \in n^2$) relation region has a relation embedding $\boldsymbol{u}_{ij}$ as well as the union box features $\boldsymbol{x}^b_{k} = \mathbf{MLP}[b_i, b_j, b_{i\cup j}, b_{i\cap j}]$. Specifically, we set the input visual features $\boldsymbol{x}^r_{k}$ to be $[\boldsymbol{\hat{x}}_{i},\boldsymbol{u}_{ij},  \boldsymbol{\hat{x}}_{j}]$. The $[\cdot]$ denotes the concatenation operation.

\textbf{Relation ResGCN.}
Methods~\cite{motif,factornet,GuZL0CL19,vctree} extract contextual information from objects' features can achieve good performance on very few frequent relations, but poor performance on the rest infrequent relations.
Our \textit{Relation ResGCN} aggregates context information of relation features and fuses the relation co-occurrence prior in a concatenation manner, which avoids the bias caused by the large probabilities of the object co-occurrence.
We now describe how the \textit{Relation ResGCN} works. 
Given a target relation which consists of its visual features $\boldsymbol{x}^r_{i}$ and spatial features $\boldsymbol{x}^b_{i} $, our \textit{Relation ResGCN} learns to collect context information from other relations to refine $\boldsymbol{x}^r_i$ and enhances its
representation ability. More formally, this message passing mechanism is defined as:
 \begin{equation}
  \boldsymbol{\hat{x}}^r_i= \boldsymbol{x}^r_i  + ReLU(  \phi_r LN([\sum_{j \neq i}e_{ij} \cdot (\boldsymbol{W}^r_V  \boldsymbol{x}^r_{j}), \boldsymbol{\hat{q}}_{i})])) ,
  \label{ResR}
\end{equation}
where $\phi_r$ is a transformation function. $LN$ is a LayerNorm layer. $`[\cdot]'$ denotes the feature concatenation operation. $\boldsymbol{W}^r_V$ is a transformation and the output of this term is a weighted sum of its neighbors' visual features. $\boldsymbol{q}_i$ are aggregated context features that depend on the relation co-occurrence prior (detailed below). The original relation features are added to ensure that the main
information of the relation is reserved and not overwhelmed
by the transformed information from others. Intuitively, the message passing in our method described by Eq. \ref{ResR} can be considered as to learn the residual representation of $\boldsymbol{x}^r_i$, which is expected to be the context information.
Thus, the final features $\boldsymbol{\hat{x}}^r_i$ include the appearance information, the relation co-occurrence prior, and the context information around it.

Concretely, the weight $\boldsymbol{e}_{ij}$ indicates the importances of its neighbors. It is calculated by matching the appearance and geometric similarity in a latent space, which is defined as:
 \begin{equation}
  e_{ij}=\frac{w^{G}_{ij}\cdot\exp(w^r_{ij})}
  { \sum_k w^{G}_{ik} \cdot \exp(w^r_{ik})}.
  \label{eq:w}
\end{equation}
The $w^{G}_{ij}$ and $w^r_{ij}$ are computed as:

 \begin{equation}
 w^G_{ij}=\frac{\max(0, \text{dot}(\boldsymbol{W}^G_r\cdot\boldsymbol{x}^b_{i}, \boldsymbol{W}^G_r\cdot\boldsymbol{x}^b_{j}) )}{\sqrt{d}},
 \label{eq:wg}
\end{equation}
 \begin{equation}
 w^r_{ij}=\frac{\text{dot}(\boldsymbol{W}^r_K\cdot\boldsymbol{x}^r_{i}, \boldsymbol{W}^r_Q\cdot\boldsymbol{x}^r_{j}) )}{\sqrt{d}}.
 \label{eq:wr}
\end{equation}
$\boldsymbol{W}^r_G$, $\boldsymbol{W}^r_K$ and $\boldsymbol{W}^r_Q$ are all transformation matrices which embed the corresponding features into a common subspace to measure how well they match. And $d$ is the channel dimension of embedded features.

For $\boldsymbol{q}_i$,  the statistical information of the relation co-occurrence is encoded. The co-occurrence probabilities of relation pairs are first counted on the training set and a matrix $\boldsymbol{M} \in \mathbb{R}^{C^r \times C^r}$ is obtained. Specifically, for two relation categories of $c$ and $c'$, we count the probability $M_{cc'}$ of the existence of  category $c$ in the presence of category $c'$.
To avoid the side effects of the frequency bias, note that we count a type of relation only once in an image. That is to say, if a relation occurs multiple times in an image, we count it only once.
Given two relation regions $i$ and $j$, we duplicate the $i$-th region $C^r$ times to obtain $C^r$ nodes, \textit{i.e.,} node $i_{1},i_{2},...,i_{C^r}$. Node $i_{c}$ denotes the correlation of the $i$-th region  with category $c$. The same process is performed for the $j$-th region. In this way, we can use $\boldsymbol{M}_{cc'}$ to correlate node $i_{c}$ and $j_{c'}$ and construct a graph for all regions. 

Specifically, the node of each region has a hidden state $\boldsymbol{q}_{ic}$.
The initial node representation $\boldsymbol{q}_{ic} $ is set to $\boldsymbol{x}^r_i$. Each note aggregates context information from its neighbors according to the graph structure, formulated as:
\begin{equation}\label{eq:aic}
\begin{split}
 \boldsymbol{q}_{ic}=\left[\sum_{j=1,j\neq i}^{n}\sum_{c'=1}^{C^r}\boldsymbol{M}_{c'c}\boldsymbol{q}_{jc'}, \sum_{j=1,j\neq i}^{n}\sum_{c'=1}^{C^r}\boldsymbol{M}_{cc'}\boldsymbol{q}_{jc'}\right].
\end{split}
\end{equation}
Eq. \ref{eq:aic} can be considered as the context features which contain the relation co-occurrence. We use the co-occurrence probabilities among nodes to measure their importance to each other and aggregate information through it.
Then, the generated final hidden state of each node is 
$\{\boldsymbol{\hat{q}}_{i1}, \boldsymbol{\hat{q}}_{i2}, \dots, \boldsymbol{\hat{q}}_{iC^r}\}$. Similar to~\cite{ggnn}, we use two fully connected layers to obtain the node-level features:
 \begin{equation}
  \boldsymbol{q}_{ic}^r=\psi_r([\boldsymbol{q}_{ic}, \boldsymbol{\hat{q}}_{ic}]).
\end{equation}
\begin{equation}
    \boldsymbol{\hat{q}}_i=\varphi_r([\boldsymbol{q}_{i1}^{r}, \boldsymbol{q}_{i2}^{r}, \dots, \boldsymbol{q}_{iC^r}^{r}]).\\
   \label{eq:object-cls}
\end{equation}
Finally, the refined features $\boldsymbol{\hat{x}}^{r}_i$ are passed to a classifier to achieve the distribution of the relation label $P^r_i$.

\noindent \textbf{Module fusion.}
For the $i$-th relation, we obtain the final probability $\hat{P_i}$ by summing the two scores with the softmax normalization:
\begin{equation}
\hat{P_i} = \text{softmax}(P^r_i + P^{r\hat{o}}_i),
\end{equation}
where $\hat{P}_i$ is the final predicted probability of relation, and $P^r_i$ and $P^{r\hat{o}}_i$ are unnormalized class logits.

\section{Experiments}
\subsection{Experiment Settings}

 \textbf{Dataset.}
 We evaluate the proposed method and existing state-of-the-art competitors on the Visual Genome (VG)~\cite{DBLP:journals/ijcv/KrishnaZGJHKCKL17} benchmark. It contains 108,073 images with tens of thousands of unique object and predicate relation categories.  However, most categories have a very limited number of instances.
 In our experiments, we follow the most commonly used data splits proposed by~\cite{XuZCF17}. The 150 most frequent object categories and the 50 most frequent predicate types are selected.
 After preprocessing, each image has 11.5 objects and 6.2 relationships on average. The dataset is split with 70\% of images as the training set and the other 30\% as the testing set, respectively. We further picked 5,000 images from the training set as the validation set for hyperparameter tuning.
 
 \textbf{Tasks.}   We followed three conventional evaluation modes: (1)
 \textbf{Scene Graph Generation} (SGGen): Given an
image, we need to detect the objects and predict their pairwise relationship classes. In particular, the object detection
needs to localize both the subject and object with at least 0.5
IoU with the ground-truth.  (2) \textbf{Scene Graph Classiﬁcation} (SGCls): Given the ground-truth object bounding boxes,
we need to predict both the object and pairwise relationship classes. 
 (3) \textbf{Predicate Classiﬁcation} (PredCls): Given the ground-truth object bounding boxes and class labels, we need to predict the visual relationship classes among all the object pairs.

\begin{figure*}[t]
  \centering
  \subfigure{
  \begin{minipage}[t]{0.5\linewidth}
  \centering
  \includegraphics[width = 1.0\columnwidth]{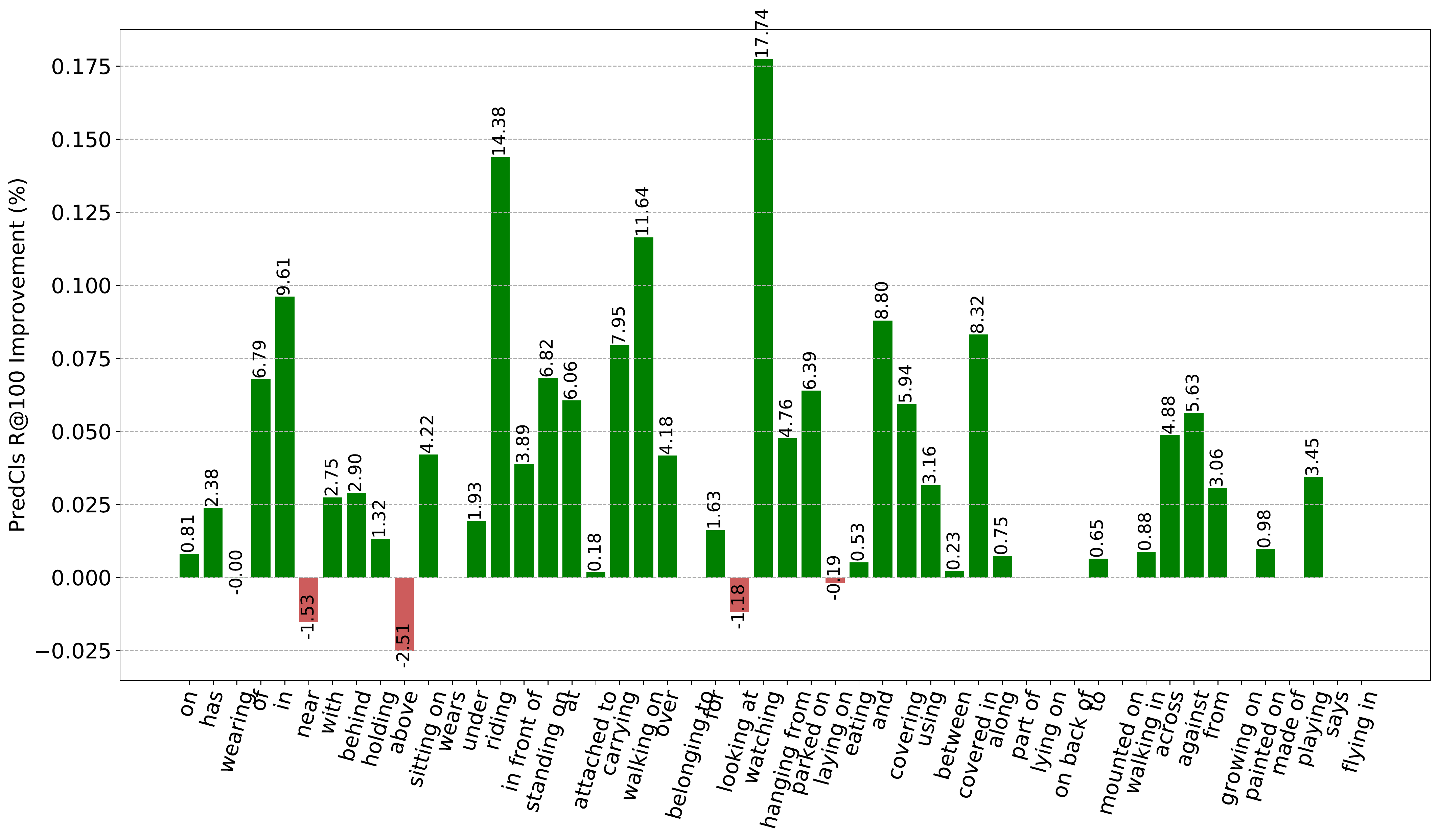}
  \end{minipage}%
  }%
  \subfigure{
  \begin{minipage}[t]{0.5\linewidth}
  \centering
  \includegraphics[width = 1.0\columnwidth]{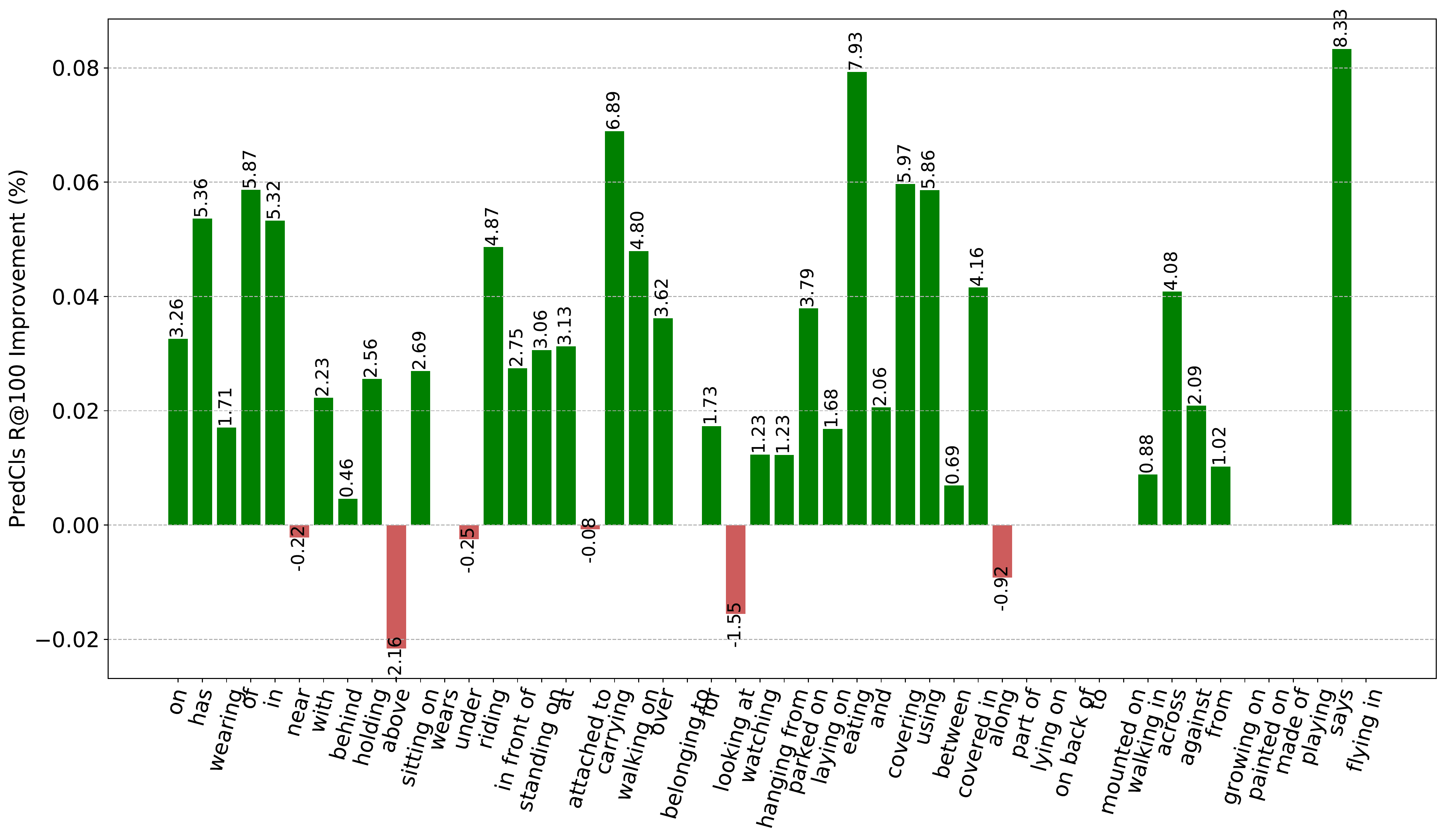}
  \end{minipage}%
  }%
  \centering
  \caption{R@100 improvement on different relations of
  our methods over the VCTREE on the predicate classification (\textbf{left}) and scene graph classification (\textbf{right}) on the VG dataset. The R@100 is computed with constraint.}
  
  \label{fig:R100}
  
  \end{figure*}

  \textbf{Evaluation metric.} 
In each task, the main evaluation metric is the average per-image recall of the top $\mathbf{K}$ subject-predicate-object triplets. The confidence score of a triplet is computed by multiplying the classification confidence scores of all the three elements.  Then it is used for ranking against the top $\mathbf{K}$ triplets in the ground-truth scene graph.  A triplet is matched if all the three elements are classified correctly, and the bounding boxes of subject and object match with an IoU of at least 0.5.
We followed the conventional Recall@K (R@K = 20,50,100) as the evaluation metrics. 
However, the most commonly used metric Recall@K is insensitive to low prediction accuracy of infrequent relations (see Fig. \ref{fig:R100}).
To this end, we introduce a balanced metric named: \textbf{mean Recall@K} (short as mR@K)~\cite{kern}. This metric computes the recall
for each predicate class separately and then averages \textbf{Recall@K} for all predicates. 
Additionally, some works omit the constraint that limits the top K triplets to only one predicate for each ordered entity pair. In this work, we report both R@K and mR@K with and without constraint respectively for comprehensive comparisons.

\begin{table*}[htbp]
  \centering
  \caption{Comparison of the R@(20,50,100) and mR@(20,50,100) in \% with and without constraint on the three tasks of the VG dataset.}
     \resizebox{0.9\textwidth}{!}{
    \begin{tabular}{c|c|c|rcccccc}
    \hline
    \multirow{2}[3]{*}{Task} & \multirow{2}[3]{*}{Metric} & \multicolumn{1}{l|}{    Graph} & \multicolumn{7}{c}{Methods} \\
\cline{4-10}          &       &    Constraint & \multicolumn{1}{c}{IMP} & IMP+  & FREQ  & SMN   & VCTREE & KERN  & Ours \\
    \hline
    \hline
    \multirow{12}[2]{*}{SGGen} & \multirow{2}[1]{*}{mR@20} & Yes   & \multicolumn{1}{c}{0.4} &   -    & 2.8   & 3.9   & 5.2   & 4.7   & \textbf{6.1} \\
          &      & No    & \multicolumn{1}{c}{-}     &  -     &  -     &     -  &    -   &  -     & -  \\
          & \multirow{2}[0]{*}{mR@50} & Yes   & \multicolumn{1}{c}{0.6} & 3.8   & 4.3   & 5.3   & 6.9   & 6.4   & \textbf{8.4} \\
          &       & No    &     \multicolumn{1}{c}{-}  & 5.4   & 5.9   & 9.3   & 12.3  & 11.7  & \textbf{13.5} \\
          & \multirow{2}[0]{*}{mR@100} & Yes   & \multicolumn{1}{c}{0.9} & 4.8   & 5.6   & 6.1   & 8.0   & 7.3   & \textbf{9.5} \\
          &       & No    &     \multicolumn{1}{c}{-}  & 8.0   & 8.9   & 12.9  & 16.5  & 16.0  & \textbf{18.6} \\
          & \multirow{2}[0]{*}{R@20} & Yes   &  \multicolumn{1}{c}{-}     & 14.6  & 17.7  & 21.4  & 22.0 &     -  & \textbf{22.1} \\
          &       & No    &     \multicolumn{1}{c}{-}  &    -   &  -     &  -     &     -  &    -   & - \\
          & \multirow{2}[0]{*}{R@50} & Yes   & \multicolumn{1}{c}{3.4} & 20.7  & 23.5  & 27.2  & 27.9 & 27.1  & \textbf{28.1} \\
          &       & No    & \multicolumn{1}{c}{9.7} & 22.0  & 25.3  & 30.5  & 31.5 & 30.9  & \textbf{31.8} \\
          & \multirow{2}[1]{*}{R@100} & Yes   & \multicolumn{1}{c}{4.2} & 24.5  & 27.6  & 30.3  & 31.3 & 29.8  & \textbf{31.5} \\
          &       & No    & \multicolumn{1}{c}{-}      & 27.4  & 30.9  & 35.8  & 37.0 & 35.8  & \textbf{37.6} \\
    \hline
    \multirow{12}[2]{*}{SGCls} & \multirow{2}[1]{*}{mR@20} & Yes   & \multicolumn{1}{c}{2.2} & 4.8   & 4.9   & 5.8   & 8.2   & 7.7   & \textbf{9.1} \\
          &       & No    &     \multicolumn{1}{c}{-}  &    -   &  -     &  -     &     -  &     -  & - \\
          & \multirow{2}[0]{*}{mR@50} & Yes   & \multicolumn{1}{c}{3.1} & 5.8   & 6.8   & 7.1   & 10.1   & 9.4   & \textbf{11.1} \\
          &       & No    &  \multicolumn{1}{c}{-}      & 12.1  & 13.5  & 15.4  & 20.6  & 19.8  & \textbf{22.8} \\
          & \multirow{2}[0]{*}{mR@100} & Yes   & \multicolumn{1}{c}{3.8} & 6.0   & 7.8   & 7.6   & 10.8   & 10.0  & \textbf{12.0} \\
          &       & No    & \multicolumn{1}{c}{-}      & 16.9  & 19.6  & 20.6  & 27.2  & 26.2  & \textbf{29.6} \\
          & \multirow{2}[0]{*}{R@20} & Yes   &   \multicolumn{1}{c}{-}    & 31.7  & 27.7  & 32.9  & 35.2 &     -  & \textbf{35.4} \\
          &       & No    &     \multicolumn{1}{c}{-}  &    -   &   -    &  -     &     -  &    -   & - \\
          & \multirow{2}[0]{*}{R@50} & Yes   & \multicolumn{1}{c}{21.7} & 34.6  & 32.4  & 35.8  & 38.1 & 36.7  & \textbf{38.3} \\
          &       & No    & \multicolumn{1}{c}{-}      & 43.4  & 40.5  & 44.5  & 47.7 & 45.9  & \textbf{47.9} \\
          & \multirow{2}[1]{*}{R@100} & Yes   & \multicolumn{1}{c}{24.4} & 35.4  & 34.0  & 36.5  & 38.8 & 37.5  & \textbf{39.1} \\
          &       & No    & \multicolumn{1}{c}{-}      & 47.2  & 43.7  & 47.7  & 50.9 & 49.0  & \textbf{51.3} \\
    \hline
    \multirow{12}[2]{*}{PredCls} & \multirow{2}[1]{*}{mR@20} & Yes   & \multicolumn{1}{c}{3.7} & 7.9   & 8.7   & 10.3  & 14.0  & 13.8  & \textbf{15.6} \\
          &       & No    & \multicolumn{1}{c}{-}      &     -  & -      & -      &     -  &    -   & - \\
          & \multirow{2}[0]{*}{mR@50} & Yes   & \multicolumn{1}{c}{6.1} & 9.8   & 13.3  & 13.3  & 17.9  & 17.7  & \textbf{19.7} \\
          &       & No    &     \multicolumn{1}{c}{-}  & 15.1  & 20.3  & 24.8  & 35.1  & 26.3  & \textbf{38.4} \\
          & \multirow{2}[0]{*}{mR@100} & Yes   & \multicolumn{1}{c}{8.0} & 10.5  & 15.8  & 14.8  & 19.4  & 19.2  & \textbf{21.5} \\
          &       & No    & \multicolumn{1}{c}{-}      & 28.9  & 37.3  & 37.9  & 47.6  & 49.0  & \textbf{51.7} \\
          & \multirow{2}[0]{*}{R@20} & Yes   &   \multicolumn{1}{c}{-}    & 52.7  & 49.4  & 58.5  & 60.1  &   -    & \textbf{60.2} \\
          &       & No    & \multicolumn{1}{c}{-} & -   & -   & -  & -  & - & - \\
          & \multirow{2}[0]{*}{R@50} & Yes   & \multicolumn{1}{c}{44.8} & 59.3  & 59.9  & 65.2  & 66.4  & 65.8  & \textbf{66.6} \\
          &       & No    &     \multicolumn{1}{c}{-}  & 75.2  & 71.3  & 81.1  & 82.9  & 81.9  & \textbf{83.2} \\
          & \multirow{2}[1]{*}{R@100} & Yes   & \multicolumn{1}{c}{53.0} & 61.3  & 64.1  & 67.1  & 68.1  & 67.6  & \textbf{68.2} \\
          &       & No    &   \multicolumn{1}{c}{-}    & 83.6  & 81.2  & 88.3  & 89.8  & 88.9  & \textbf{90.2} \\
    \hline
    \end{tabular}%
    }
  \label{tab:main}%
\end{table*}%

\subsection{Implementation Details}
\noindent \textbf{Object detector.} 
We adopt Faster-RCNN with the VGG backbone to detect object bounding boxes and extract RoI features. For a fair comparison, we use the same set of parameters as~\cite{motif}. Moreover, the anchor boxes size and aspect ratio are adjusted similar to YOLO-9000~\cite{yolo9000}, and the RoIPooling layer is replaced with a RoIAlign layer~\cite{maskrcnn}.

\noindent \textbf{Training details. }
We freeze the layers before the ROIAlign layer and optimize the whole framework with the sum of the object cross-entropy loss and the relationship cross-entropy loss. The batch size and initial learning rate are set to 8 and 0.00001, respectively. 
Our model is optimized by the Adam algorithm with the momentums of 0.9 and 0.999. We divide the learning rate by 10 when the recall of the validation set plateaus. For SGDet, we follow the post-processing step in~\cite{motif} and~\cite{vctree} for a fair comparison. Since the number of all possible relationship pairs are huge (\textit{e.g.}, 64 objects leads to more than 4,000 pairs), we follow \cite{motif, vctree, kern} that only consider the relationships between two objects with overlapped bounding boxes,
which reduces the number of object pairs to around 1,000.
After predicting the object class probabilities for each RoI, we
use a per-class NMS to select the RoI class and its corresponding class-specific offsets from Faster-RCNN. The IoU threshold in NMS is set to 0.3 in our experiments.

\subsection{Comparisons with State-of-the-art Methods}
\textbf{Competing methods.} In this part, we compare our proposed method with several existing state-of-the-art methods. 
We group these methods into three groups: (1) FREQuency baseline (FREQ)~\cite{motif} is an independent inference model,
which predicts the classes of objects and relations independently.
(2) (IMP)~\cite{XuZCF17}, its improved version by using a better detector (IMP+)~\cite{XuZCF17}, and Stacked Motif Networks (SMN)~\cite{motif} are joint inference models, which adopt message passing to encode the context. (3) Visual Context Tree model (VC-TREE)~\cite{vctree} and Knowledge-Embedded Routing Network (KERN)~\cite{kern} are also joint inference models. And they attempt to solve the imbalance issue.

\smallskip
\noindent
\textbf{Comparison of overall performance.} 
The quantitative results are reported in Table~\ref{tab:main}. We can observe that our model achieves the state-of-the-art performance under all the Recall and mean Recall
evaluation metrics. Specifically, compared to the famous baseline SMN, our method achieves relative improvements of 55.7\%, 57.9\%, and 45.2\% in mR@100 on SGGen, SGCls, and PredCls, respectively.
Compared to VC-TREE, it is worth nothing that our method can especially improve the performance of three tasks significantly in mean Recall@100 (\emph{i.e.,} 28.4\%, 11.1\%, and 10.8\% relative improvements on SGGen, SGCls, and PredCls, respectively). 
For the Recall metric, despite our training process doesn't skew on frequent categories, our method still achieves improved performance over other state-of-the-art methods. The improvements in mean Recall and Recall meet our architecture design, where our dual ResGCN does not merely learn the class distribution bias, but improves the performance on infrequent predicates and really aids in alleviating the imbalance issue in scene graph generation. 
For SGDet, the improvements are not as significant as SGCls, the
reason may come from the imperfect and noisy detected bounding boxes.

\smallskip
\noindent
\textbf{Detailed comparisons and analyses.} Compared to the state-of-the-art methods, we find that the improvements are significant in both mR@K and R@K.
However, previous methods which achieve good overall performance in Recall show extremely low Recall on semantically informative predicates.
The frustrating fact is that a model that only predicts the top 10 most frequent relations can even achieve a score of 90\% under the Recall metric. Because the Recall metric is dominated by the performance of these most frequent relations. As the model can not be effectively evaluated by Recall, mean Recall is a more important metric than overall Recall especially when the class distribution of the database is severely skewed.
To give a comprehensive analysis of this phenomenon, we present the improvement in R@100 for each relation on PredCls and SGCls in Fig.~\ref{fig:R100}. Note that the x-axis labels are in descending order based on their sample proportions in the VG dataset. As depicted in Fig.~\ref{fig:R100}, our model achieves obvious improvements in almost all relations, especially for these infrequent relations.  
 The VC-TREE is greatly inclined to the top 5 most frequent relations such as `on' and `has', which is the reason why its Recall on SGCls is comparable to ours. However, VC-TREE achieves quite poor performance for these infrequent relations~(\emph{e.g.,} `\emph{parked on}' and `\emph{walking on}').
Specifically, more than 30\%  of `\emph{parked on}', `\emph{walking on}', and `\emph{sitting on}' predicates are misclassified to the less informative predicate 
`\emph{on}' by VC-TREE. While our model only fails on 15\% of that. Such results indicate that our model can generate more fine-grained relation features.



\subsection{Ablation Experiments}
\textbf{1) Ablation studies.} 
To prove the effectiveness of our proposed methods, we
conduct two ablation studies. Results of the ablation studies
are summarized in Table ~\ref{tab:ablationstudy} and Table ~\ref{tab:N}, respectively.

\noindent\textbf{Effectiveness of the proposed components.}
The core of our method is the explicit incorporation of the \textit{Relation ResGCN}, the \textit{Object ResGCN}, and the relation prior.
To better demonstrate the effectiveness of each component, we drop each part with leaving the other components unchanged. Then, we retrain the model in the same way. 
(i) We first analyze the model structures. `O-ResGCN' and `R-ResGCN' denote the \textit{Object ResGCN} branch and the \textit{Relation ResGCN} branch, respectively.
The detailed descriptions of them  are provided in Section \ref{section:3.1} and \ref{section:3.2}. Table~\ref{tab:ablationstudy} clearly shows that the performance improves consistently when all the components are used together.
In particular, the `R-ResGCN' is the most important for the performance of infrequent relations, because dropping the `R-ResGCN' branch leads to a dramatic decrease in the performance under mean Recall.  
Besides, removing the `O-ResGCN'  will lead to a clear performance drop in Recall, especially for SGCLs task and SGGen task. This is because disabling the object-level context modeling critically hurt the inference of object labels. For PredCls task, we can see that the `O-ResGCN' also plays a critical role in inferring relationships.
Thus, such obvious performance drop in mean Recall and Recall clearly indicates that the two separate ResGCNs are complementary. Fusing the `O-ResGCN' and the `R-ResGCN` not only allows our model to learn the bias of the dataset but also significantly improves the performance on infrequent predicates.
(ii)  In addition, we analyze the contributions of the priors.  Removing the relation prior and training the dual ResGCN directly will also lead to a dramatic decrease in mean Recall while not hurt the performance much in Recall, which proves that the combination of ResGCN and relation prior does benefit the infrequent relation prediction.

\begin{figure*}[t]
    \centering
    \includegraphics[width = 0.99\textwidth]{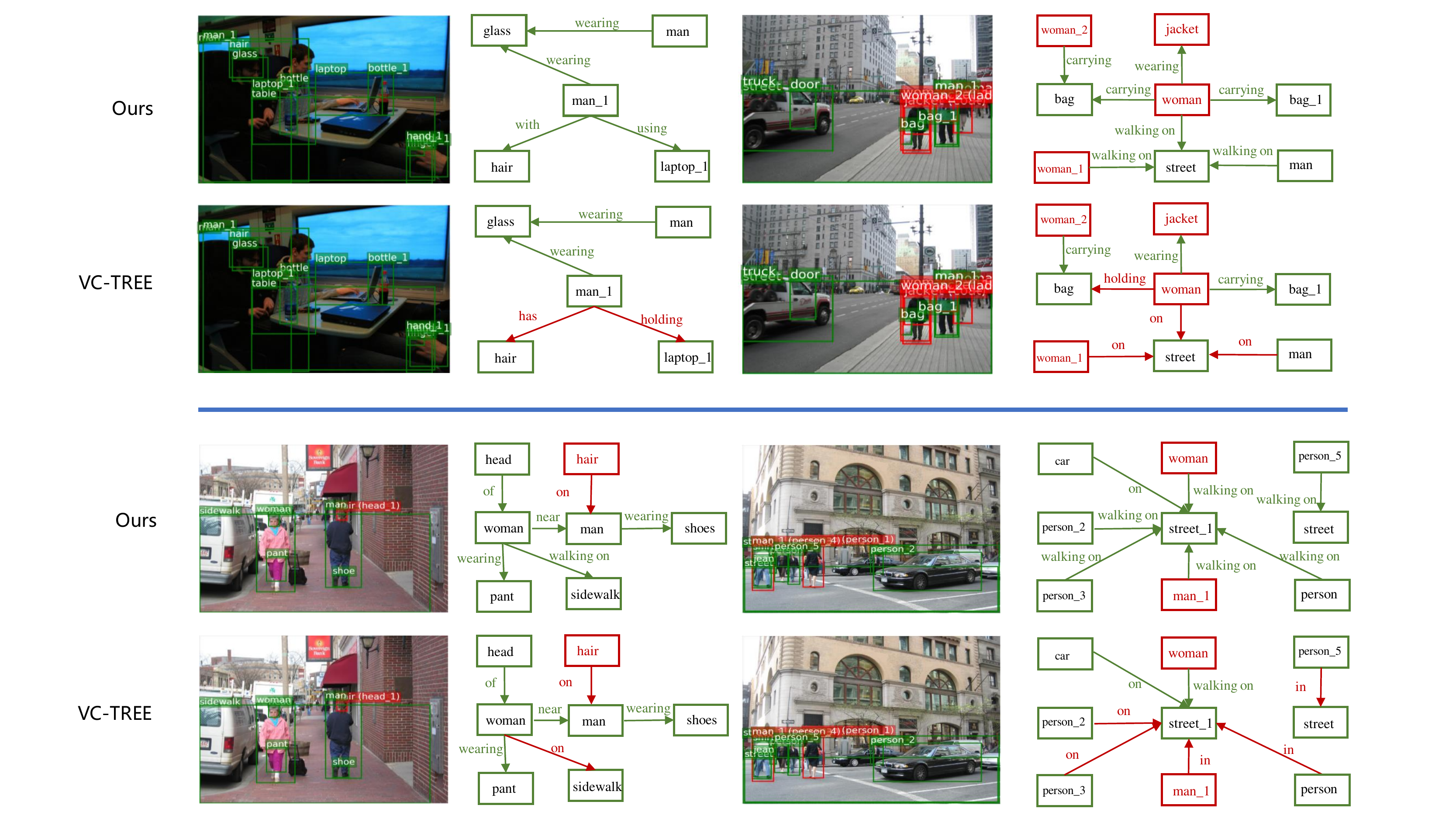}
    \caption{
    Qualitative results showing comparisons between VC-TREE and our method in SGCls. 
Green boxes are bounding boxes with the ground-truth, and red boxes are
ground-truth with no match. Green edges are true positive predicted by each model at the mR@100 setting, and red edges are false negatives.}
    \label{fig:Qualitative_results}
\end{figure*}

\noindent\textbf{Number of contextual  coefficients $N$.} We conduct additional analyses on the number of contextual  coefficients $N$ (details in Section \ref{section:3.1}).  The contextual coefficient represents a  possible importance weight from one object node to another. As depicted in Table~\ref{tab:N}, stacking more number of contextual coefficients steadily improves the performance. The improvement saturates at $N=4$, where $0.9\%$ gain in mR@100 is achieved on the SGCls task.

 \begin{table*}[htbp]
  \centering
  \caption{Ablation studies on our proposed components.}
  \resizebox{0.95\textwidth}{!}{
    \begin{tabular}{c|cccc|cccc|cccc}
    \toprule
    \multirow{2}[2]{*}{Model} & \multicolumn{4}{c|}{PredCls}  & \multicolumn{4}{c|}{SGCls}    & \multicolumn{4}{c}{SGGen} \\
          & R@50  & R@100 & mR@50 & mR@100 & R@50  & R@100 & mR@50 & mR@100 & R@50  & R@100 & mR@50 & mR@100 \\
    \hline
    Ours full model & 66.6  & 68.2  & 19.7  & 21.5  & 38.3  & 39.1  & 11.1  & 12.0    & 28.1  & 31.5  & 8.4   & 9.5 \\
    Ours w/o O-ResGCN & 65.9  & 67.6  & 19.5  & 21.1  & 36.8  & 37.9  & 8.9   & 10.1  & 26.9  & 29.7  & 6.7   & 7.4 \\
    Ours w/o R-ResGCN & 66.4  & 67.9  & 18.3  & 19.9  & 38.1  & 38.7  & 10.2  & 11.1  & 27.7  & 30.9  & 7.9   & 8.8 \\
    Ours w/o Relation Prior & 66.3  & 68.1  & 18.9  & 20.5  & 38.2  & 39.0    & 10.8  & 11.5  & 27.9  & 31.2  & 8.2   & 9.1 \\
    \hline
    \end{tabular}%
    }
    
  \label{tab:ablationstudy}%
\end{table*}%

 \begin{table*}[htbp]
  \centering
  \caption{Ablation studies on the Number of contextual  coefficients.}
  \resizebox{0.95\textwidth}{!}{
    \begin{tabular}{c|cccc|cccc|cccc}
    \toprule
    \multirow{2}[2]{*}{Model} & \multicolumn{4}{c|}{PredCls}  & \multicolumn{4}{c|}{SGCls}    & \multicolumn{4}{c}{SGGen} \\
          & R@50  & R@100 & mR@50 & mR@100 & R@50  & R@100 & mR@50 & mR@100 & R@50  & R@100 & mR@50 & mR@100 \\
    \hline
    $N$ = 1 & 66.1  & 67.7  & 19.2  & 20.9  & 37.8  & 38.4  & 10.6  & 11.1    & 27.4  & 30.8  & 7.6   & 8.7 \\
    $N$ = 2 & 66.3  & 68.0  & 19.5  & 21.3  & 38.1  & 38.9  & 10.8   & 11.6  & 27.8  & 31.1  & 8.1   & 9.1 \\
    $N$ = 4 & 66.6  & 68.2  & 19.7  & 21.5  & 38.3  & 39.1  & 11.1  & 12.0    & 28.1  & 31.5  & 8.4   & 9.5 \\
    $N$ = 8 & 66.3  & 67.9  & 19.4  & 21.2  & 38.2  & 39.0    & 10.9  & 11.7  & 28.0  & 31.3  & 8.2   & 9.3 \\
    \hline
    \end{tabular}%
    }
    
  \label{tab:N}%
\end{table*}%
 
\noindent \textbf{2) Cost-sensitive learning.}
 Except for the models that we have discussed before, we also investigate a cost-sensitive learning method: \textbf{Online Hard Example Mining (OHEM)}~\cite{ohem}. This method is introduced to alleviate the class imbalance in the training process. It is widely used in recognition and detection tasks. 
 In our task, the OHEM is proposed to force networks to focus on hard (and so more valuable) relations during training process.
The loss function is defined as:
\begin{equation}
\begin{split}
    \mathcal{L}(\boldsymbol{P}, \boldsymbol{\hat{R}}) =& - \frac{\sum^{l}_{i = 1}\sum^{C^r}_{j=1}\textbf{1}\left\{
    y_i = j\;  and\;  P_{ij} < t
    \right\} \log P_{ij}}{\sum^{l}_{i = 1}\sum^{C^r}_{j=1}\textbf{1}
    \left\{
    y_i = j\;  and\;  P_{ij} < t
    \right\},
    }
\end{split}
\end{equation}
where $l$ is the number of relations in a graph.
$\boldsymbol{P}$ and $ \boldsymbol{\hat{R}} \in \mathbb{R}^{l \times C^r}$ are the matrices of relation scores and labels, respectively. $y_i = \arg\max \boldsymbol{r}_i$ is the ground-truth label of the $i$-th relation.
$P_{ij}$ denotes the predicted probability of the $i$-th relation belonging to the relation category $j$.
$t \in (0, 1] $ is a threshold. Here $\textbf{1}\{ \cdot \}$ equals one when the condition inside holds, and otherwise equals zero. 
In practice, we hope that there should be at least a reasonable
number of relations kept per graph.
Thus, we first sort the loss in descending order. Then, the top $\tau$ of the losses will be selected.
We set $\tau$ to 0.7 in our experiments.  We replace our original cross entropy loss with the OHEM loss, and the final best results on PredCls are:
\textbf{``R@100: 68.0\%", ``mR@100: 20.7\%"}, which are even worse than our reported results \textbf{``R@100: 68.2\%", ``mR@100: 21.5\%"}.
Besides, the result is very sensitive to the hyperparameters which are very difficult to tune, while our method incorporates the prior into the structure without additional hyperparameters.

\begin{figure*}[t]
    \centering
    \includegraphics[width = 0.99\textwidth]{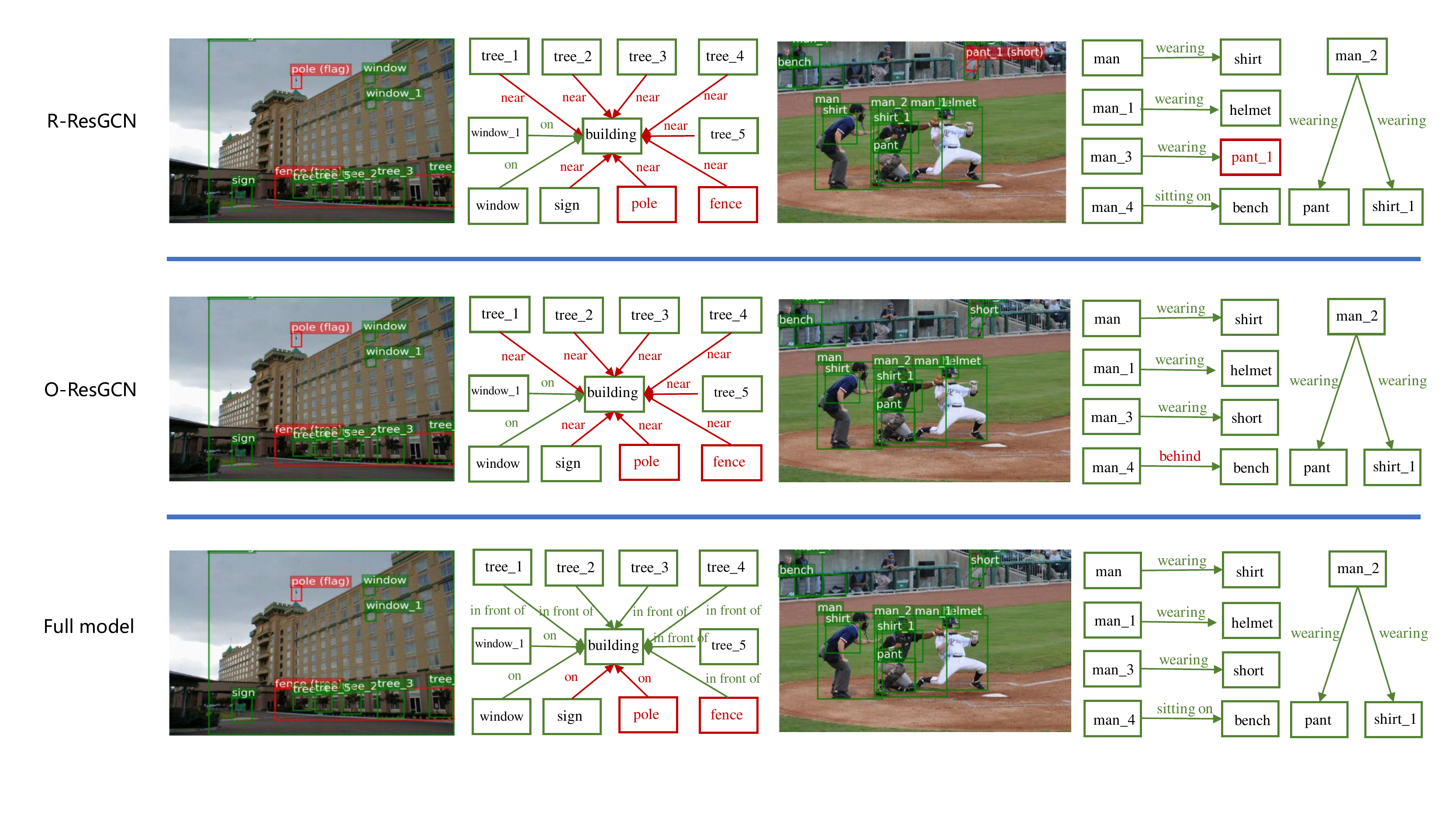}
    \caption{Qualitative results showing comparisons among \textit{Relation ResGCN} branch, \textit{Object ResGCN} and our full model
in SGCls. Green boxes are bounding boxes with the ground-truth, and red boxes are
ground-truth with no match. Green edges are true positive predicted by each model at the mR@100 setting, and red edges are false negatives.}
    \label{fig:Qualitative_results_2}
\end{figure*}

\section{Qualitative Results} \label{sec:3}
Here we present some qualitative results of scene graph generation in Fig.~\ref{fig:Qualitative_results} and Fig.~\ref{fig:Qualitative_results_2}. 
In Fig.~\ref{fig:Qualitative_results}, the upper part illustrates the results of our method.  The lower part illustrates the results of the state-of-the-art method VC-TREE~\cite{vctree}. Green boxes are bounding boxes with the ground-truth and red boxes are the ground-truth with no match. Green edges are truely predicted by each model under the mR@100 setting and red edges are false predictions.
As shown in figures, both of our method and VC-TREE successfully predict the frequent predicates ( \textit{e.g.,} \emph{`on', `wearing', `near', and etc..}) while VC-TREE is more likely to fail on some infrequent predicates such as \emph{`walking on', `using', `holding', and etc.}.
Those infrequent relations normally carry more semantic information. Thus, our method tends to predict informative predicates rather than trivial ones, which is more capable to understand complicated scenes. In Fig.~\ref{fig:Qualitative_results_2}, we present the generated scene graphs of the \textit{Relation ResGCN} branch, the \textit{Object ResGCN} branch, and our full model. As shown, the combination of the \textit{Relation ResGCN} and the \textit{Object ResGCN} does benefit the prediction of infrequent relationships.

\section{Conclusion}
In this paper, we propose a novel model to address the class imbalance in scene graph generation. 
Unlike the network structures of previous methods, our {\textit{dual ResGCN}} encodes the relation-level context and the object-level context separately through the \textit{Relation ResGCN} and the \textit{Object ResGCN}, respectively.
The two networks are complementary to each other to maintain the performance on the prediction of frequent relations and improve the performance of infrequent relations. 
Besides the novel network, we also propose to incorporate prior knowledge into the context aggregation process to alleviate the side effects of the class imbalance. 
Extensive experiments and analyses demonstrate the superiority of the proposed method over competing methods, especially the improvements in the prediction of infrequent relations.

\ifCLASSOPTIONcaptionsoff
  \newpage
\fi

\bibliographystyle{IEEEtran}
\bibliography{egbib}

\begin{IEEEbiography}[{\includegraphics[width=1in,height=1.25in,clip,keepaspectratio]{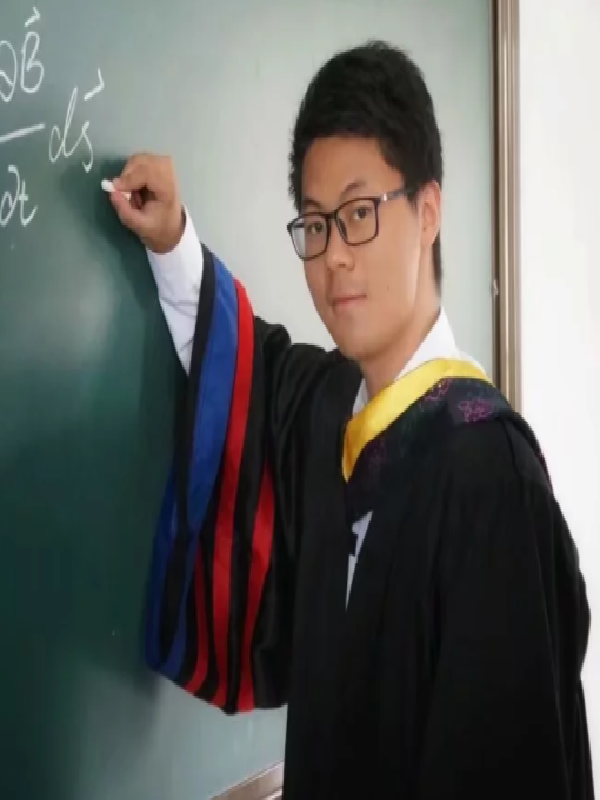}}]{Jingyi Zhang} is currently a master student in School of Computer Science and Engineering, University of Electronic Science and Technology of China.
His current research interests include multimedia and computer
vision.
\end{IEEEbiography}

\begin{IEEEbiography}
	[{\includegraphics[width=1in,height=1.25in,clip,keepaspectratio]{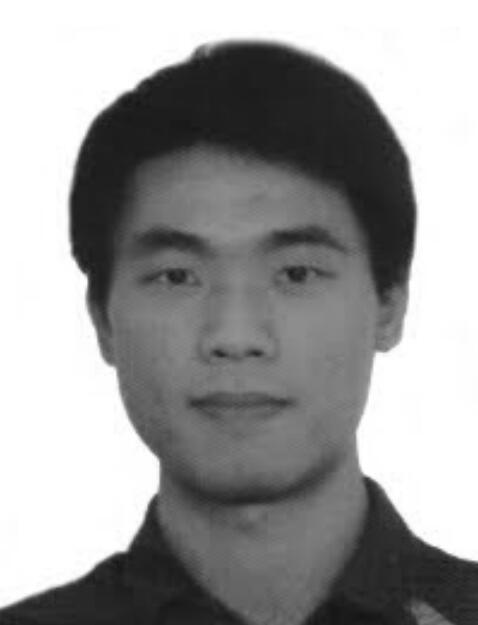}}]
{Yong Zhang}	
received the Ph.D. degree 	in pattern recognition and intelligent systems from the Institute of Automation, Chinese Academy of Sciences in 2018. From 2015 to 2017, he was a Visiting Scholar with the Rensselaer Polytechnic Institute. He is currently with the Tencent AI Lab. His research interests include computer vision, machine learning, and probabilistic graphical models.
\end{IEEEbiography}

\begin{IEEEbiography}
	[{\includegraphics[width=1in,height=1.5in,clip,keepaspectratio]{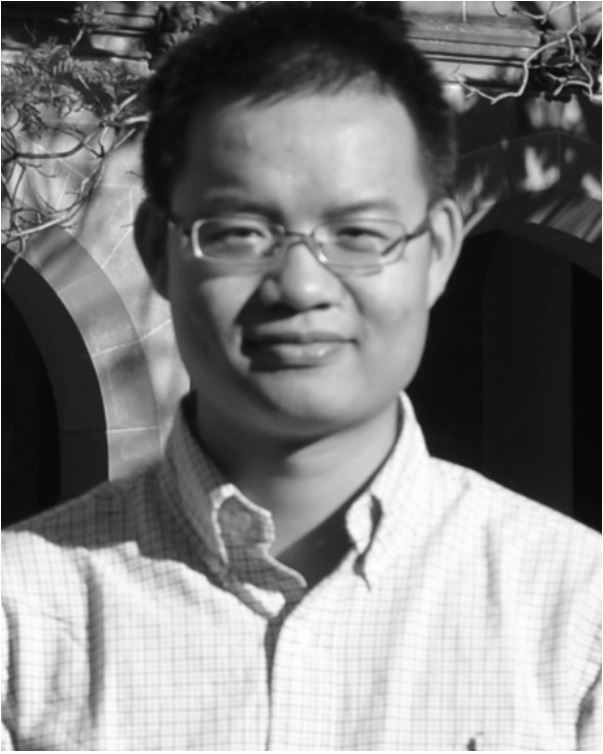}}]
{Baoyuan Wu} (Member, IEEE) received the Ph.D. degree from the National Laboratory of Pattern Recognition, Chinese Academy of Sciences, in 2014. He is currently an Associate Professor with School of Data Science, The Chinese University of Hong Kong, Shenzhen. He is also the director of the Secure Computing Lab of Big Data, Shenzhen Research Institute of Big Data. He was Senior and Principal Researcher with Tencent AI Lab, from November 2016 to August 2020. He held a postdoctoral position at the IVUL Lab, KAUST, working with Prof. B. Ghanem, from August 2014 to November 2016. His research interests are machine learning, computer vision, optimization, as well as AI security and privacy.
\end{IEEEbiography}

\begin{IEEEbiography}
	[{\includegraphics[width=1in,height=1.5in,clip,keepaspectratio]{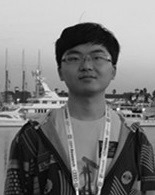}}]
	{Yanbo Fan}
	is currently a Senior Researcher at Tencent AI Lab. He received his Ph.D. degree from Institute of Automation, Chinese Academy of Sciences (CASIA), Beijing, China, in 2018, and his B.S. degree in Computer Science and Technology from Hunan University in 2013. His research interests are computer vision and machine learning.
\end{IEEEbiography}

\begin{IEEEbiography}[{\includegraphics[width=1in,height=1.5in,clip,keepaspectratio]{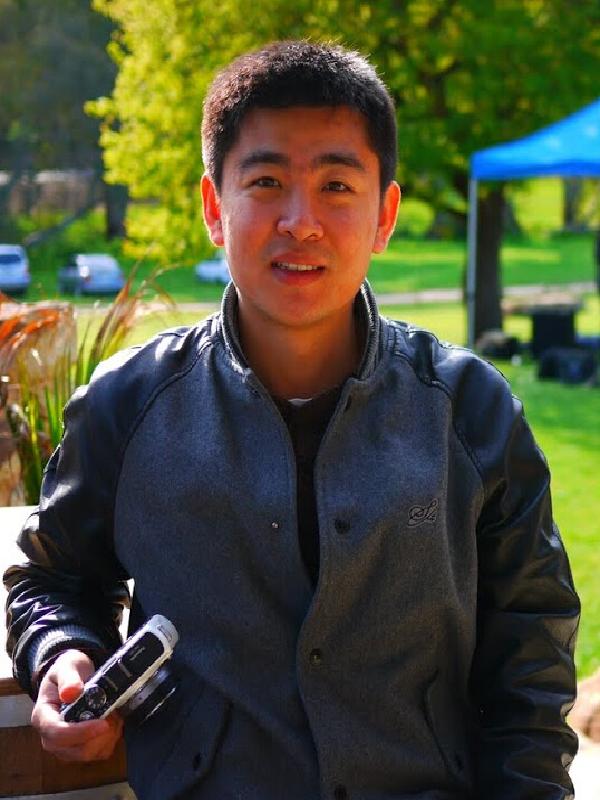}}]{Fumin Shen} (Member, IEEE)
received his Bachelor degree at 2007 and PhD degree at 2014 from Shandong University and Nanjing University of  Science and Technology, China, respectively. Now he is  with  School of Computer Science and Engineering, University of Electronic Science and Technology of China. His major research interests include computer vision and machine learning. He was the recipient of the Best Paper Award Honorable Mention at ACM SIGIR 2016, ACM SIGIR 2017 and the  World's FIRST 10K Best Paper Award - Platinum Award at IEEE ICME 2017.
\end{IEEEbiography}

\begin{IEEEbiography}[{\includegraphics[width=1in,height=1.5in,clip,keepaspectratio]{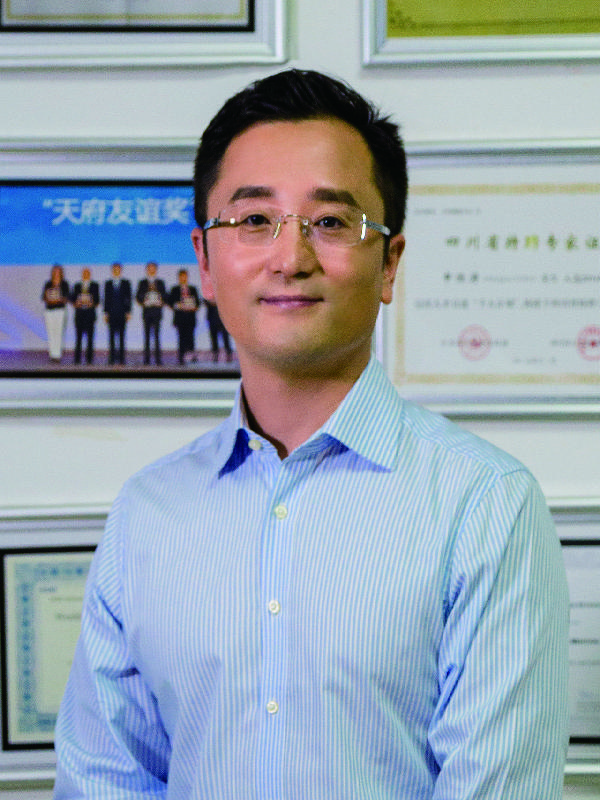}}]{Heng Tao Shen}(Senior Member, IEEE) is a Professor of National ``Thousand Talents Plan'' and the director of Center for Future Media at the University of Electronic Science and Technology of China (UESTC). He obtained his BSc with 1st class Honours and PhD from Department of Computer Science, National University of Singapore (NUS) in 2000 and 2004 respectively. He then joined the University of Queensland (UQ) as a Lecturer, Senior Lecturer, Reader, and became a Professor in late 2011. His research interests mainly include Multimedia Search, Computer Vision, and Big Data Management on spatial, temporal, and multimedia databases. He has published over 150 peer-reviewed papers, most of which are in prestigious international venues of interests. For his outstanding research contributions, he received the Chris Wallace Award in 2010 conferred by Computing Research and Education Association, Australasia, and the Future Fellowship from Australia Research Council in 2012. He is an Associate Editor of IEEE Transactions on Knowledge and Data Engineering, and has organized ICDE 2013 as Local Organization Co-Chair, and ACM Multimedia 2015 as Program Committee Co-Chair. 
\end{IEEEbiography}

\end{document}